\documentclass[runningheads]{llncs}
\usepackage[T1]{fontenc}
\usepackage{graphicx}
\usepackage[ruled,linesnumbered]{algorithm2e}
\usepackage{float}
\usepackage{amsmath}
\usepackage{amssymb}

\usepackage{cancel}
\usepackage{placeins}

\usepackage{array}

\usepackage{CJK}
\usepackage{setspace}
\usepackage{multicol} 
\usepackage{multirow}
\usepackage{booktabs}  
\usepackage{threeparttable}
\usepackage{algpseudocode} 

\usepackage{subcaption}
\usepackage{color}

\begin{document}
\title{Enhancing Emotion Recognition in Conversation through Emotional Cross-Modal Fusion and Inter-class Contrastive Learning}

\author{Haoxiang Shi\inst{1,2\dagger}, Xulong Zhang\inst{1\dagger}, Ning Cheng\inst{1}, Yong Zhang\inst{1}, Jun Yu\inst{2}, Jing Xiao\inst{1} and Jianzong Wang\inst{1}\thanks{Corresponding author: Jianzong Wang (jzwang@188.com)\\$\dagger$ Equal Contribution}}
\authorrunning{H. Shi et al.}
\titlerunning{Enhancing Emotion Recognition in Conversation}
%
\institute{Ping An Technology (Shenzhen) Co., Ltd., Shenzhen, China\and
University of Science and Technology of China, Hefei, China}
\maketitle              
\begin{abstract}
The purpose of emotion recognition in conversation (ERC) is to identify the emotion category of an utterance based on contextual information. Previous ERC methods relied on simple connections for cross-modal fusion and ignored the information differences between modalities, resulting in the model being unable to focus on modality-specific emotional information. At the same time, the shared information between modalities was not processed to generate emotions. Information redundancy problem. To overcome these limitations, we propose a cross-modal fusion emotion prediction network based on vector connections. The network mainly includes two stages: the multi-modal feature fusion stage based on connection vectors and the emotion classification stage based on fused features. Furthermore, we design a supervised inter-class contrastive learning module based on emotion labels. Experimental results confirm the effectiveness of the proposed method, demonstrating excellent performance on the IEMOCAP and MELD datasets.

\end{abstract}
\section{Introduction}
Emotions are a crucial element in human communication \cite{mim,zy2}. In the realm of dialogues between humans and machines, the ability to discern variations in a user's emotional state is of significant importance. The incorporation of emotion recognition techniques \cite{ec241,zy1} allows for the development of more empathetic and sensitive intelligent systems. This, in turn, can lead to a more pleasant user experience and a more efficient exchange between humans and machines.

Previous emotion recognition (ERC) models in conversations only used single-modal information (like text and audio) for recognition \cite{ec241,ec242} , such as simply applying speech emotion recognition (SER) \cite{ser24} to conversations, which is useful to some extent. In daily conversations, the information we are exposed to is diverse, and the use of multiple modalities, such as text, audio, and vision, greatly improves the inference of human emotions because each modality captures different aspects of emotional expression \cite{ec243}. Therefore, the effective fusion of multi-modal information is crucial to improving the accuracy and comprehensiveness of predictions. Many works have introduced multi-modality into emotion recognition, which has greatly improved the performance of emotion recognition. For example, Zhao et al. \cite{scfa} proposed a speaker-perceived multi-modal emotion recognition network, and Kim et al. \cite{concat23} integrated multiple audio features and text features to enhance emotion recognition. 

However, these works still have the following problems with the utilization of multi-modal information: 1) Previous ERC methods use the same modeling approaches to process text and audio data \cite{fus232}, ignoring differences between modalities \cite{fus231}. 2) During modal fusion, previous models tend to directly connect the two \cite{concat24,concat2}, causing information redundancy, as the text modality primarily contains content information, while the audio modality captures more prosodic information \cite{aia}. Additionally, this approach overlooks the interaction between modalities. Information between different modalities is often complementary, and a reasonable combination can improve performance.

To address these issues, we propose a multi-step fusion model based on joint vectors to identify emotions in multi-party conversations. Specifically, the model consists of two stages. In the first stage, we extract the mel-spectrogram of the audio and text embeddings to represent the information of each modality. In the second stage, based on the dual-modality information input, we constructed a multi-modal fusion module using trainable joint vectors and separately encoded the two modalities through pre-trained large language and visual models. Multi-modal information interaction is facilitated by joint vectors, thereby enabling information interaction while ensuring the independence of each modality. The emotion recognition task is then completed based on the final generated fusion features. Additionally, to address the problem of imbalanced emotional category samples, we proposed an inter-class contrastive learning module. Using emotional sample labels, we supervised the separation of inter-class samples while simultaneously reducing the distance between intra-class samples to improve the representational ability of multi-modal features. The main contributions of this work are summarized as follows:

\begin{itemize}
\item We propose a multi-modal emotion recognition model utilizing joint vectors. We employ pre-trained models for modeling and joint vectors for modal interaction, ensuring the independence of each modality.
\item We propose an inter-class contrastive learning module to alleviate the problem of imbalanced emotional sample categories via supervised learning based on emotional labels.
\item We conduct experiments on two widely-used datasets, experimental results demonstrate that our model achieves leading performance.
\end{itemize}

\begin{figure*}[ht]
\begin{minipage}[b]{1.0\linewidth}
  \centering
  \centerline{\includegraphics[width=12cm]{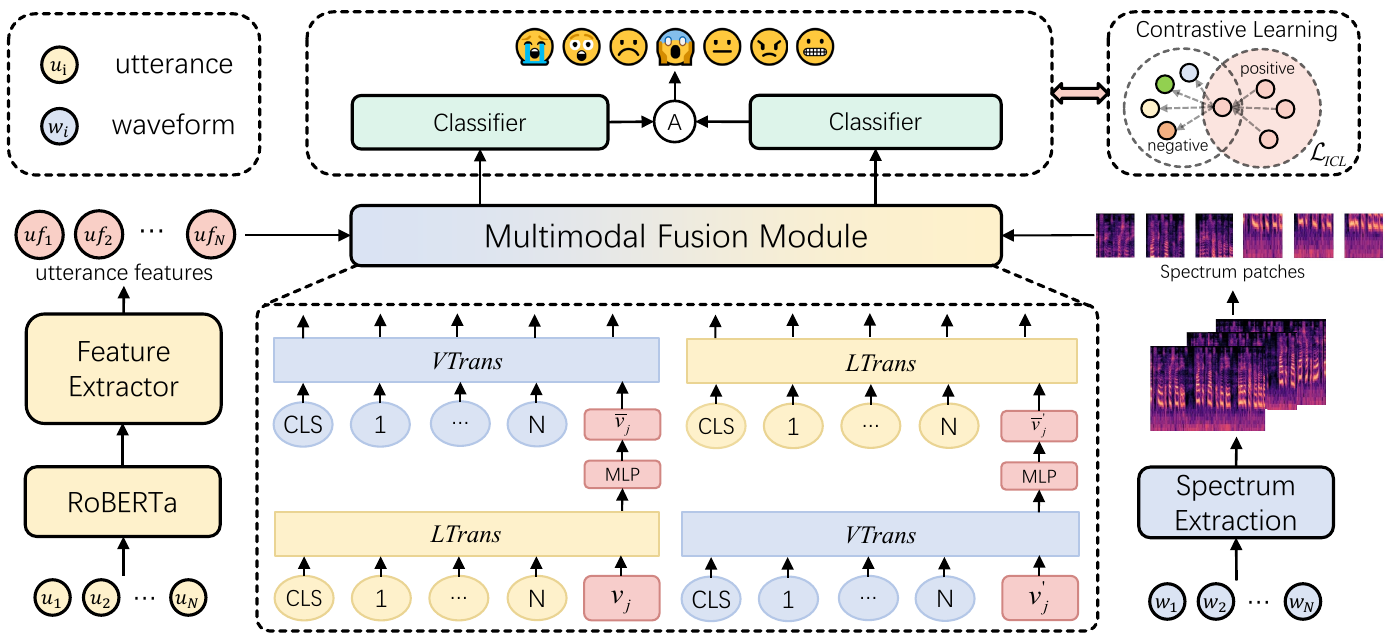}}
  \caption{The architecture of our model. The left and right sides represent text and audio inputs, respectively. By combining text and audio features, the cross-modal joint vector $v_j$ is used for multi-modal information fusion to extract the fused feature representation, and finally, perform emotion classification. Simultaneously, different colors in the contrast learning module represent different emotional category samples.}
  \label{arc}
\end{minipage}
\end{figure*}

\section{Methods}
\subsection{Task Definition}

\textbf{Emotion Recognition in Conversation (ERC).} Within a multi-modal discourse denoted by $\mathbb{D}$, we encompass both textual and auditory components, organized as $\mathbb{D} = \{(u_1, s_1), (u_2, s_2), ..., (u_N, s_N)\}$. Each tuple $(u_i,s_i)$ corresponds to the $i^{th}$ distinct utterance attributed to a speaker within the interaction, with $N$ signifying the number of utterances. The goal of Emotion Recognition in Conversations is to identify the emotion category label $E=\{e_1, e_2, ..., e_N\}$ associated with the dialogue.

\subsection{Overview}
The model is mainly divided into three parts: the multi-modal fusion part, the emotion recognition part, and the contrastive learning part, as shown in Fig.~\ref{arc}. Specifically, in the multi-modal fusion module, we design cross-modal joint vectors to integrate information from dual modalities and generate the final multi-modal fusion representation through multiple stacked large model-based fusion modules. We then perform emotion recognition in conversations based on the generated multi-modal fusion representation. In addition, to further improve the multi-modal emotional representation capabilities, we design a contrastive learning module to use emotional labels for inter-class optimization on the generated multi-modal representations, thereby enhancing the recognition effect. Some modules in these components are introduced below:
\begin{itemize}
  \item \textbf{RoBERTa encoder} operates as an encoder for the extraction of semantic content from text data. For this purpose, we select the output from the 12-th layer of the large-RoBERTa \cite{roberta} model as the text embedding $\mathcal{F}_t$.
  \item \textbf{Feature extractor} is used to extract deeper text features and consists of two layers of stacked transformer \cite{trans} blocks.
  \item \textbf{Spectrum extraction} executes a short-time Fourier transform (STFT) on the input audio signal, subsequently translating the frequency axis to the mel scale, thereby obtaining the mel-spectrogram.
\end{itemize}

\subsection{Joint-based Multi-modal Fusion}
As mel-spectrograms contain rich prosodic information, as detailed in \cite{prosody}.To explore the potential of audio frequency domain data and combine it with text features, we developed a joint-based fusion module (JFM), as illustrated in Fig.~\ref{arc}. First, in order to deal with the text and spectral domain data, we leverage pre-trained models from the Language \cite{bert} and Visual \cite{vit} Transformer domains, denoted by $LTrans$ and $VTrans$, respectively. For processing the mel-spectrogram, the initial step involves segmenting it into multiple patches to align with the transformer layer's input requirements. This segmentation enables us to derive the feature embeddings $\mathcal{F}_{m}$ via a linear projection method, as referenced in \cite{vit}. Following this, we introduce a trainable joint vector, initialized in accordance with the approach outlined in \cite{pmf}, and termed $\mathbf{v}_{j}$. This vector aids in the seamless blending of multi-modal fusion specifics. Our JFM contains $N$ joint-based fusion (JF) blocks. Taking the $l^{th}$ layer as an example, we concatenate $\mathbf{v}_{j}^l$ with the input $\mathbf{\mathcal{F}}_{m \to t}^l$ (the output of the $(l-1)^{th}$ JF layer, where $\mathbf{\mathcal{F}}_{m \to t}^0=\mathbf{\mathcal{F}}_{m}$), and then feed them into the visual transformer layer $VTrans^l$ to building a joint:
\begin{equation}
\begin{aligned}
(\mathbf{\hat{\mathcal{F}}}_{m\to t}^l\oplus \mathbf{\hat{v}}_{j}^l)= VTrans^l(\mathcal{F}_{m\to t}^l\oplus \mathbf{v}_{j}^l)
\end{aligned}
\end{equation}
where $\oplus$ stands for the concatenate operation. Then, the obtained joint $\mathbf{\hat{v}}_{j}^l$ with spectral domain information is mapped to the feature space of the $\mathbf{\mathcal{F}}_{t\to m}^l$ (where $\mathbf{\mathcal{F}}_{t \to m}^0=\mathbf{\mathcal{F}}_{t}$) through an MLP layer, and concatenated with the embedded $\mathcal{F}_{t \to m}^l$. These are then fed into language Transformer layer $LTrans^l$ to obtain the one-sided fusion features $\mathbf{\mathcal{F}}_{m\to t}^{l+1}$:
\begin{equation}
\begin{aligned}
(\mathcal{F}_{m \to t}^{l + 1} \oplus \mathbf{V}_j^l) = LTrans^l(\mathcal{F}_{t \to m}^l \oplus \bar{\mathbf{v}}_{j}^l)
\end{aligned}
\end{equation}
where $\bar{\mathbf{v}}_{j}^l= MLP(\mathbf{\hat{v}}_{j}^l)$. Similarly, as for the other-sided fusion features $\mathbf{\mathcal{F}}_{t \to m}^{l+1}$, we adapt symmetric operations:
\begin{equation}
\begin{aligned}
(\mathbf{\hat{\mathcal{F}}}_{t \to m}^l\oplus \mathbf{\hat{v}}^{\prime l}_{j})= LTrans^{\prime l}(\mathcal{F}_{t \to m}^l\oplus \mathbf{v}^{\prime l}_{j})
\end{aligned}
\end{equation}
\begin{equation}
\begin{aligned}
(\mathbf{\mathcal{F}}_{t\to m}^{l+1} \oplus \mathbf{V}_{j}^{\prime l}) = VTrans^{\prime l}(\mathbf{\mathcal{F}}_{m\to t}^l \oplus \bar{\mathbf{v}}_{j}^{\prime l})
\end{aligned}
\end{equation}
Where $\mathbf{v}^{\prime l}_{j}$ represents different joint vectors from $\mathbf{v}_{j}^l$, $LTrans^{\prime l}$ and $VTrans^{\prime l}$ share the same pre-trained weights as $LTrans^{l}$ and $VTrans^{l}$.

\subsection{Emotion Recognition}
After acquiring the bilateral fusion features, we take the output representation of CLS token of $\mathbf{\mathcal{F}}_{m\to t}^N$ and $\mathbf{\mathcal{F}}_{t\to m}^N$ as inputs to two distinct linear classifiers. The logits are then averaged to determine the final recognition $\hat{r}_i$ emotions, respectively. The classifier consists of a fully connected layer and a softmax layer. Finally, our classification loss is formulated as:
\begin{equation}
\begin{aligned}
\mathcal{L}_{ERC} =-\sum_{i} r_i \log(\hat{r}_i)
\end{aligned}
\end{equation}
where $r_i$ represent emotion labels. 

\subsection{Inter-class Contrastive Learning}
In the ERC task, due to the imbalance in the number of samples of each category in the dataset, some emotional categories may be ignored in the process of calculating the loss. Therefore, we construct inter-class contrastive learning (ICL) to enable the model to focus on the differences between samples of different emotional categories \cite{cl}. Using label information, all samples with the same emotional label in the same batch are regarded as positive samples, and emotional samples of other categories are regarded as negative samples, thereby shortening the distance between samples of the same category and more clearly distinguishing them from samples of other categories. Assuming there are K samples in a batch, firstly, to comprehensively consider the cross-fusion features output by the multi-modal fusion module, we first concatenate the two fusion features and obtain feature $\mathcal{F}$:
\begin{equation}
    \begin{aligned}
        \mathcal{F} = [\mathbf{\mathcal{F}}_{m\to t}^N \oplus \mathbf{\mathcal{F}}_{t\to m}^N]
    \end{aligned}
\end{equation}
then we calculate the inter-class contrastive learning loss $\mathcal{L}_{ICL}$ using the following method:
 \begin{equation}
     \begin{aligned}
        \mathcal{L}_{ICL} = \sum_{i \in I}\frac{-1}{N_{P(i)}}\sum_{p \in P(i)}log \frac{exp((\mathcal{F}_i \cdot \mathcal{F}_p)/\tau)}{\sum_{j=0,j\ne i}^{K}exp(\mathcal{F}_i \cdot \mathcal{F}_j / \tau)}
       \end{aligned}
 \end{equation}
where $i \in I=\{1,2,\ldots, K\}$ represents the index corresponding to samples within a batch, $\tau$ signifies a positive real-valued temperature parameter that is utilized to regulate the spacing among the samples. $P(i)$ refers to the collection of samples that share the same emotional category as the $i^{th}$ sample, $N_{P(i)}$ denotes the number of samples in $P(i)$.

\section{Experiments}
\subsection{Datasets and Baselines}
We have performed experimental evaluations utilizing two distinct datasets: the IEMOCAP \cite{iemocap} and the MELD \cite{meld}. The IEMOCAP is a renowned audiovisual database employed extensively in the realm of affective computing, encompassing around 12 hours of recorded material. It is specifically curated for interactions involving two participants. The IEMOCAP dataset segments each dialogue into distinct speech units, applying continuous labeling within the Valence-Arousal emotional spectrum and categorical labeling across various emotions such as $anger$, $happiness$, $sadness$, and $neutrality$. On the other hand, the MELD dataset is a leading resource for tasks involving the analysis of emotions expressed by multiple speakers. It comprises over 1400 conversations and approximately 13000 speech instances excerpted from the television show $Friends$. The emotional labeling within this dataset encompasses a broader spectrum with seven distinct emotion labels, which include $neutral$, $happiness$, $surprise$, $sadness$, $anger$, $disgust$, and $fear$.

To demonstrate the effectiveness of our model in both single-channel and multi-channel contexts, we have chosen the following benchmark models:

\begin{itemize}
\item \textbf{SCFA}(2023) \cite{scfa} advances an innovative architecture that integrates cross-modal information with a focus on speaker-specific characteristics to enhance the emotion recognition.
\item \textbf{MultiEMO}(2023) \cite{memo} proposed a new attention-based correlation-aware multi-modal fusion framework.
\end{itemize}

\subsection{Implementation Details}
In our research, we leverage the RoBERTa model, which has been pre-trained as referenced in \cite{roberta}, to derive features with 768 dimensions from textual data. For constructing the feature extractor, we have employed a stack of two transformer layers, with each layer comprising 8 attention heads and generating embeddings of 1024 dimensions. For the audio modality, we process the source audio to obtain an 80-dimensional mel-spectrogram. Within the JFM, we configure the parameter $N$ to be 2 and define the length of the $\mathbf{v}_{j}$ as 4. For an in-depth examination of these parameters, please refer to the parameter analysis in Section 3.6. The entire neural network was optimized using the Adam, with a batch size set to 32 and a learning rate of 0.0001.

\subsection{Results Analysis}

\begin{figure}[t]
	\centering
	\begin{subfigure}{0.395\linewidth}
		\centering
		\includegraphics[width=\linewidth]{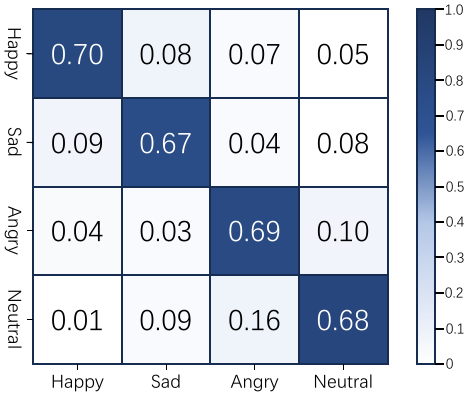}
		\caption{IEMOCAP}
        \label{f3a}		
	\end{subfigure}
	\begin{subfigure}{0.4\linewidth}
		\centering
		\includegraphics[width=\linewidth]{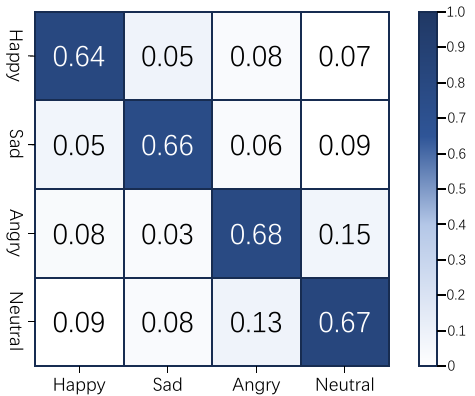}
		\caption{MELD}
        \label{f3b}
	\end{subfigure}
\caption{Accuracy confusion matrix on two datasets.}
\label{f3}
\end{figure}

\begin{table*}[t]
    \centering
    \renewcommand{\arraystretch}{1.1}
	\caption{Comparison of performance. `-' means there is no relevant data in the original paper, `$*$' indicates that our results are statistically significant under the t-test ($p < 0.05$) relative to the comparison model. `T': Text, `A': Audio.}
        \begin{tabular}{p{2cm}<{\centering} p{2.5cm}<{\centering} p{0.2cm}<{\centering} p{1.5cm}<{\centering}       p{1.5cm}<{\centering} p{0.2cm}<{\centering} p{1.5cm}<{\centering} p{1.5cm}<{\centering}}
            \hline
            \multicolumn{2}{c}{Dataset}&&\multicolumn{2}{c}{IEMOCAP}&&\multicolumn{2}{c}{MELD}\\
            \hline
            Modality & Method && Acc$\uparrow$ & W-F1$\uparrow$ && Acc$\uparrow$ & W-F1$\uparrow$\\
            \hline
            \multirow{3}*{T} 
            & SCFA \cite{scfa} && 63.82 & 62.89 && 59.32 & 57.76\\
            & MultiEMO \cite{memo} && - & \textbf{64.48} && - & \textbf{61.23}\\
            & Ours && \textbf{64.83} & 64.23 && \textbf{62.82*} & 60.68 \\ 
            \hline
            \multirow{3}*{A} 
            & SCFA \cite{scfa} && 49.24 & 48.66 && 47.37 & 44.18\\
            & MultiEMO \cite{memo} && - & 38.89 && - & 33.55 \\
            & Ours && \textbf{50.37*} & \textbf{49.76*} && \textbf{48.97*} & \textbf{45.42*}\\ 
            \hline
            \multirow{3}*{T$+$A} 
            & SCFA \cite{scfa} && 67.91 & 66.42 && 64.86 & 63.69\\
            & MultiEMO \cite{memo} && - & \textbf{69.18} && - & 64.21 \\
            & Ours && \textbf{68.77} & 68.66&& \textbf{65.62} & \textbf{64.73} \\ 
            \hline
        \end{tabular}
        \label{t2}
\end{table*}
To validate the model's performance, we conducted experiments on the two datasets with accuracy (Acc) and weighted F1 (W-F1) metrics, as shown in Table~\ref{t2}. Our model achieves comparable performance across all three modal combinations. Specifically, for example on MELD, our model outperforms SCFA's Acc by 4.50$\%$ under the single text modality, slightly lower than MultiEMO by 0.55\% In the audio modality, our model demonstrates improved performance compared to SCFA by 1.24\% and significantly outperforms MultiEMO by 11.87\%. Notably, MultiEMO’s inferior performance may be due to its lack of audio-specific feature modeling. Finally, in multi-modal experiments, our model exhibits comparable performance to SCFA and MultiEMO, surpassing both works on the MELD dataset. Additionally, to demonstrate the recognition of specific emotion categories, we constructed a confusion matrix based on four emotion categories, as shown in Fig.~\ref{f3}.

\subsection{Ablation Study}
We conducted ablation experiments by removing the JFM and ICL modules to assess the effectiveness of our proposed module, as depicted in Table~\ref{ab}. Taking the results from the IEMOCAP dataset as an example, initially, we removed the JFM module, which involved directly utilizing RoBERTa and feature extraction to output text features and mel-spectrum for late fusion. We employed a separate ViT \cite{vit} to encode the mel-spectrogram. The results indicated that Acc and W-F1 decreased by 4.67\% and 4.53\% respectively, demonstrating the significant enhancement of modal fusion efficacy by the multi-modal fusion module. Additionally, we attempted to remove the joint vector $\mathbf{v}_j$, disrupting the interaction between the two modalities. This resulted in Acc and W-F1 decreasing by 2.08\% and 1.76\% respectively, indicating that the proposed joint vector effectively facilitates information exchange between modalities. Finally, we removed the ICL module, resulting in Acc and W-F1 decreasing by 1.32\% and 1.04\% respectively. This indicates that while the multi-modal fusion features incorporate some emotional information, removing the ICL module diminishes the model's ability to resist sample imbalances, thereby reducing its robustness.

\begin{table}[t]
\renewcommand{\arraystretch}{1.1}
\centering
\caption{The results of ablation study. ``w/o'' stands for without various modules.}
\begin{tabular}{p{2.5cm}<{\centering} p{0.2cm}<{\centering} p{1.5cm}<{\centering}       p{1.5cm}<{\centering} p{0.2cm}<{\centering} p{1.5cm}<{\centering} p{1.5cm}<{\centering}}
    \hline
    Dataset&&\multicolumn{2}{c}{IEMOCAP}&&\multicolumn{2}{c}{MELD}\\
    \hline
    Method && Acc$\uparrow$ & W-F1$\uparrow$ && Acc$\uparrow$ & W-F1$\uparrow$\\
    \hline
     Ours && /&/&&/&/\\
    w/o JFM && -4.67 & -4.53 && -5.64 & -5.01\\  			
    w/o $\mathbf{v}_j$ && -2.08 & -1.76 && -1.93 & -1.64\\
    w/o ICL && -1.32 & -1.04 && -0.97 & -0.84\\ 
    \hline
\end{tabular}
\label{ab}
\end{table}

\begin{table}[t]
\renewcommand{\arraystretch}{1.2}
\centering
\caption{The results of different fusion methods.}
\begin{tabular}{p{2.5cm}<{\centering} p{0.2cm}<{\centering} p{1.5cm}<{\centering}       p{1.5cm}<{\centering} p{0.2cm}<{\centering} p{1.5cm}<{\centering} p{1.5cm}<{\centering}}
    \hline
    Dataset&&\multicolumn{2}{c}{IEMOCAP}&&\multicolumn{2}{c}{MELD}\\
    \hline
    Method && Acc$\uparrow$ & W-F1$\uparrow$ && Acc$\uparrow$ & W-F1$\uparrow$\\
    \hline
     JFM(ours) && /&/&&/&/\\
    CFA \cite{scfa} && -1.45 & -1.23 && -1.69 & -1.56\\
    Concatenate && -2.65 & -2.73 && -3.21 & -2.98\\
    \hline
\end{tabular}
\label{abfus}
\end{table}

\subsection{Fusion Method Comparison} 
In addition, to validate the effectiveness of JFM as the fusion module, we conducted experiments based on two different fusion methods: cross-modal fusion attention (CFA) \cite{scfa} and feature concatenation. Like JFM, we employed a pre-trained ViT \cite{vit} to extract features from the mel-spectrogram for subsequent modal interaction. The experimental results, presented in Table~\ref{abfus}, indicate that replacing JFM with CFA led to a decrease in W-F1 on IEMOCAP by 1.23\%. This suggests that JFM demonstrates superior modal information fusion capability compared to CFA, and the joint vector effectively facilitates inter-modal information transmission. Additionally, directly concatenating the mel-spectrogram features with the text features $\mathcal{F}_{t}$ and inputting them into the classifier resulted in a decrease in accuracy by 2.73\%. This means that direct concatenation has limited effect on modal interaction.
\begin{figure}[t]
	\centering
	\begin{subfigure}{0.43\linewidth}
		\centering
		\includegraphics[width=\linewidth]{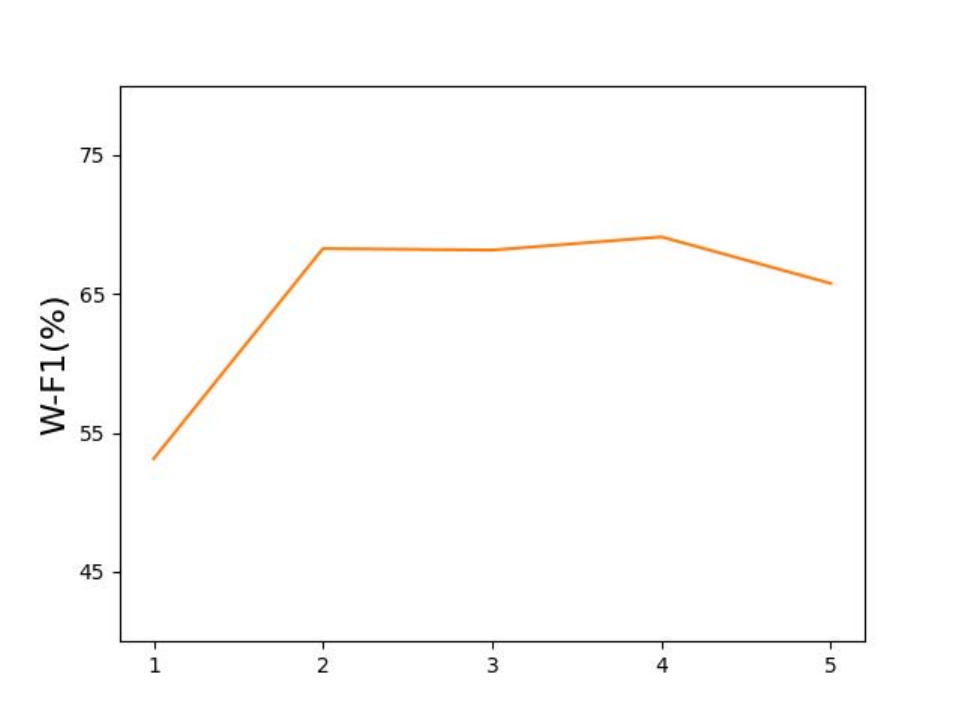}
		\caption{numbers of JF blocks.}
        \label{f4a}		
	\end{subfigure}
	\begin{subfigure}{0.43\linewidth}
		\centering
		\includegraphics[width=\linewidth]{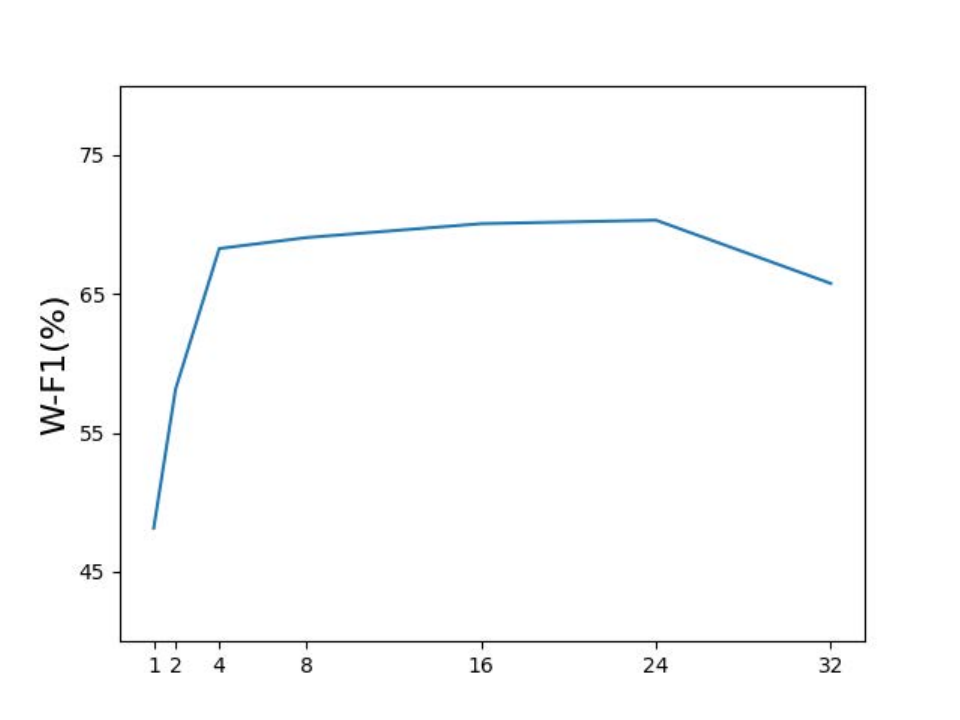}
		\caption{lengths of joint vector.}
        \label{f4b}
	\end{subfigure}
\caption{Model performance under different numbers of JF blocks and different lengths of joint vector.}
\label{f4}
\end{figure}

\subsection{Parameter Analysis} 
We conducted experiments to assess the sensitivity of our model to hyperparameters, focusing mainly on two parameters in multi-modal fusion: the number of JF blocks $N$ and the length of joint vector $\mathbf{v}_{j}$. We varied $N$ from (1, 2, 3, 4, 5) and the length of $\mathbf{v}_{j}$ from (1, 2, 4, 8, 16, 24, 32). The experimental results on the IEMOCAP dataset, depicted in Fig.~\ref{f4}, illustrate the impact of these parameters. In Fig.~\ref{f4}(a), we observe that increasing the number of JF modules generally improves accuracy indicators, suggesting enhanced extraction of emotional information through stacked JF modules. However, the effectiveness of improvement becomes limited as $N > 2$, and an increase in $N$ is accompanied by higher memory usage. Conversely, in Fig.~\ref{f4}(b), it is evident that the performance initially improves and then declines with the increase in the length of the joint vector $\mathbf{v}_{j}$. This phenomenon may arise from the incorporation of redundant modal information in excessively long joint vectors, leading to decreased accuracy. Similar to the effect of $N$, augmenting the vector length also results in higher memory consumption. Therefore, to strike a balance between memory usage and model performance, we opted for 2 layers of JF blocks for stacking, with the length of the joint vector set to 4.

\section{Conclusion}
This paper presents a multi-modal fusion model for emotion recognition based on joint vectors. Text features and audio features are individually modeled using pre-trained large models to preserve the unique information within each modality, and joint vectors facilitate modal interaction. Additionally, we introduce inter-class contrastive learning based on emotional labels to further enhance the clustering of fused features. Experimental results validate the effectiveness of each proposed module and illustrate the overall superiority of the model performance.

\section{Acknowledgement}
This paper is supported by the Key Research and Development Program of Guangdong Province under grant No.2021B0101400003. Corresponding author is Jianzong Wang (jzwang@188.com) from Ping An Technology (Shenzhen) Co., Ltd..

\bibliographystyle{splncs04}
\bibliography{z_ref.bib}

\begin{thebibliography}{10}
\providecommand{\url}[1]{\texttt{#1}}
\providecommand{\urlprefix}{URL }
\providecommand{\doi}[1]{https://doi.org/#1}

\bibitem{iemocap}
Busso, C., Bulut, M., Lee, C.C., Kazemzadeh, A., Mower, E., Kim, S., Chang, J.N., Lee, S., Narayanan, S.S.: Iemocap: Interactive emotional dyadic motion capture database. Language resources and evaluation  \textbf{42},  335--359 (2008)

\bibitem{concat24}
Chen, F., Shao, J., Zhu, A., Ouyang, D., Liu, X., Shen, H.T.: Modeling hierarchical uncertainty for multimodal emotion recognition in conversation. {IEEE} Trans. Cybern.  \textbf{54}(1),  187--198 (2024)

\bibitem{bert}
Devlin, J., Chang, M., Lee, K., Toutanova, K.: {BERT:} pre-training of deep bidirectional transformers for language understanding. In: Proceedings of the 2019 Conference of the North American Chapter of the Association for Computational Linguistics: Human Language Technologies, {NAACL-HLT} 2019. pp. 4171--4186 (2019)

\bibitem{vit}
Dosovitskiy, A., Beyer, L., Kolesnikov, A., Weissenborn, D., Zhai, X., Unterthiner, T., Dehghani, M., Minderer, M., Heigold, G., Gelly, S., Uszkoreit, J., Houlsby, N.: An image is worth 16x16 words: Transformers for image recognition at scale. In: 9th International Conference on Learning Representations, {ICLR} 2021 (2021)

\bibitem{ec241}
George, S.M., Ilyas, P.M.: A review on speech emotion recognition: {A} survey, recent advances, challenges, and the influence of noise. Neurocomputing  \textbf{568},  127015 (2024)

\bibitem{cl}
Khosla, P., Teterwak, P., Wang, C., Sarna, A., Tian, Y., Isola, P., Maschinot, A., Liu, C., Krishnan, D.: Supervised contrastive learning. CoRR  \textbf{abs/2004.11362} (2020)

\bibitem{concat23}
Kim, K., Cho, N.: {Focus-attention-enhanced Crossmodal Transformer with Metric Learning for Multimodal Speech Emotion Recognition}. In: Interspeech 2023, 24th Annual Conference of the International Speech Communication Association. pp. 2673--2677 (2023)

\bibitem{ser24}
Leem, S., Fulford, D., Onnela, J., Gard, D., Busso, C.: Selective acoustic feature enhancement for speech emotion recognition with noisy speech. IEEE/ACM Transactions on Audio, Speech, and Language Processing  \textbf{32},  917--929 (2024)

\bibitem{ec242}
Li, X., Liu, J., Xie, Y., Gong, P., Zhang, X., He, H.: {MAGDRA:} {A} multi-modal attention graph network with dynamic routing-by-agreement for multi-label emotion recognition. Knowledge-Based Systems  \textbf{283},  111126 (2024)

\bibitem{pmf}
Li, Y., Quan, R., Zhu, L., Yang, Y.: Efficient multimodal fusion via interactive prompting. In: {IEEE/CVF} Conference on Computer Vision and Pattern Recognition, {CVPR} 2023. pp. 2604--2613 (2023)

\bibitem{roberta}
Liu, Y., Ott, M., Goyal, N., Du, J., Joshi, M., Chen, D., Levy, O., Lewis, M., Zettlemoyer, L., Stoyanov, V.: Roberta: A robustly optimized bert pretraining approach. arXiv preprint arXiv:1907.11692  (2019)

\bibitem{ec243}
Peng, C., Chen, K., Shou, L., Chen, G.: {CARAT:} contrastive feature reconstruction and aggregation for multi-modal multi-label emotion recognition. In: Thirty-Eighth {AAAI} Conference on Artificial Intelligence, {AAAI} 2024. pp. 14581--14589 (2024)

\bibitem{meld}
Poria, S., Hazarika, D., Majumder, N., Naik, G., Cambria, E., Mihalcea, R.: {MELD:} {A} multimodal multi-party dataset for emotion recognition in conversations. In: Proceedings of the 57th Conference of the Association for Computational Linguistics, {ACL} 2019. pp. 527--536 (2019)

\bibitem{memo}
Shi, T., Huang, S.: Multiemo: An attention-based correlation-aware multimodal fusion framework for emotion recognition in conversations. In: Proceedings of the 61st Annual Meeting of the Association for Computational Linguistics, {ACL} 2023. pp. 14752--14766 (2023)

\bibitem{zy2}
Tang, H., Zhang, X., Wang, J., Cheng, N., Xiao, J.: Qi-tts: Questioning intonation control for emotional speech synthesis. In: IEEE International Conference on Acoustics, Speech and Signal Processing {ICASSP} 2023. pp.~1--5 (2023)

\bibitem{trans}
Vaswani, A., Shazeer, N., Parmar, N., Uszkoreit, J., Jones, L., Gomez, A.N., Kaiser, L., Polosukhin, I.: Attention is all you need. In: Advances in Neural Information Processing Systems 30: Annual Conference on Neural Information Processing Systems 2017. pp. 5998--6008 (2017)

\bibitem{concat2}
Wang, P., Zeng, S., Chen, J., Fan, L., Chen, M., Wu, Y., He, X.: {Leveraging Label Information for Multimodal Emotion Recognition}. In: Interspeech 2023, 24th Annual Conference of the International Speech Communication Association. pp. 4219--4223 (2023)

\bibitem{fus231}
Wei, J., Hu, G., Tuan, L.A., Yang, X., Zhu, W.: Multi-scale receptive field graph model for emotion recognition in conversations. In: IEEE International Conference on Acoustics, Speech and Signal Processing {ICASSP} 2023. pp.~1--5 (2023)

\bibitem{prosody}
Yang, S., Tantrawenith, M., Zhuang, H., Wu, Z., Sun, A., Wang, J., Cheng, N., Tang, H., Zhao, X., Wang, J., Meng, H.: Speech representation disentanglement with adversarial mutual information learning for one-shot voice conversion. In: Interspeech 2022, 23rd Annual Conference of the International Speech Communication Association. pp. 2553--2557 (2022)

\bibitem{mim}
Zhang, T., Chen, Z., Zhong, M., Qian, T.: Mimicking the thinking process for emotion recognition in conversation with prompts and paraphrasing. In: Proceedings of the Thirty-Second International Joint Conference on Artificial Intelligence, {IJCAI} 2023. pp. 6299--6307 (2023)

\bibitem{aia}
Zhang, T., Li, S., Chen, B., Yuan, H., Chen, C.L.P.: Aia-net: Adaptive interactive attention network for text-audio emotion recognition. IEEE Transactions on Cybernetics  \textbf{53}(12),  7659--7671 (2023)

\bibitem{scfa}
Zhao, H., Li, B., Zhang, Z.: {Speaker-aware Cross-modal Fusion Architecture for Conversational Emotion Recognition}. In: Interspeech 2023, 24th Annual Conference of the International Speech Communication Association. pp. 2718--2722 (2023)

\bibitem{fus232}
Zhao, Z., Wang, Y., Wang, Y.: Knowledge-aware bayesian co-attention for multimodal emotion recognition. In: IEEE International Conference on Acoustics, Speech and Signal Processing {ICASSP} 2023. pp.~1--5 (2023)

\bibitem{zy1}
Zhu, K., Zhang, X., Wang, J., Cheng, N., Xiao, J.: Improving eeg-based emotion recognition by fusing time-frequency and spatial representations. In: IEEE International Conference on Acoustics, Speech and Signal Processing {ICASSP} 2023. pp.~1--5 (2023)

\end{thebibliography}
\end{document}